\documentclass[runningheads]{llncs}
\usepackage[T1]{fontenc}
\usepackage{graphicx}
\usepackage{booktabs}
\usepackage{multirow}
\usepackage{algorithm}
\usepackage{algorithmic}
\usepackage{amsfonts}
\usepackage{amsmath}
\usepackage{subcaption}
\usepackage{xcolor}
\usepackage{tabularx}
\usepackage{color}

\begin{document}
\title{Have You Been Convinced? A Neuro-symbolic Argumentation Reasoning Approach to Detecting Persuasiveness in ChangeMyView Discussions}
\titlerunning{Persuasiveness Detection via Neuro-symbolic Argument Reasoning}
%
\author{Siddharth Bhargava\inst{1,2}\orcidID{0000-0002-2682-8557} \and
Sara Tonelli\inst{1}\orcidID{0000-0001-8010-6689} \and
Patricia Martín-Rodilla\inst{3}\orcidID{0000-0002-1540-883X}}
\authorrunning{S. Bhargava et al.}
%
\institute{Fondazione Bruno Kessler, Trento, Italy\\
\email{\{sbhargava,satonelli\}@fbk.eu} \and
Universidade da Coruña, Coruña, Spain \and
IEGPS-CSIC, Spanish National Research Council Spain\\
\email{p.m.rodilla@iegps.csic.es}}

\maketitle  
\begin{abstract}
Argument structures represent how individual arguments interact within a discourse. However, leveraging these structures to infer meaningful knowledge remains challenging, as it requires a formal representation of how argumentative influence propagates across the structure. Moreover, large language models (LLMs) often produce opaque predictions, limiting trustworthiness in sensitive domains such as politics, law, and healthcare. We propose Neuro-symbolic Argumentation Reasoning System (NARS) for performing persuasiveness detection directly from online discussions. The neural component employs retrieval-augmented iterative LLM reasoning to extract bipolar argument structures from long discussions, while the symbolic component applies gradual argumentation semantics to propagate argumentative influence until convergence.

Using the Webis-CMV-20 corpus, we construct CMV2BAS, a dataset of long, multi-threaded ChangeMyView discussions annotated with LLM-derived bipolar argument structures and labeled with delta events (view change vs.\ no change). We investigate whether gradual argumentation semantics can effectively detect persuasiveness, whether augmenting them with argument quality and graph-based centrality improves performance, and whether symbolic reasoning remains necessary in the presence of modern LLMs. NARS achieves F1=0.57 and AUC=0.56 under its best configuration, outperforming zero-shot Qwen3.6:27B and GPT-oss:20B, which fail to surpass random performance. We identify support-dominant discussions and long argumentative chains as key failure modes limiting attack influence propagation. Our work demonstrates the feasibility of operationalizing formal gradual argumentation semantics over noisy real-world discourse for scalable macro-level persuasion analysis, while highlighting key challenges and directions for future research in neuro-symbolic argumentation reasoning. We release both CMV2BAS dataset and the NARS framework for public use.\footnote{\label{fn:repo}Our open-source project repository: \url{https://github.com/The-obsrvr/NARS-Pipeline}}

\end{abstract}

\section{Introduction}
Argument mining (AM), a core research area under Knowledge Representation and Reasoning, studies how arguments within a discourse are proposed, debated, and ultimately accepted. In dialogical settings, this becomes particularly important, as multiple participants take turns to present and challenge arguments with the goal of reaching a resolution or consensus~\cite{visserArgumentation2016US2020,morio.fujita_2018b}. Analyzing argumentative interactions has applications in identifying key arguments, determining speaker stance, predicting discussion outcomes, and supporting explainable reasoning through interpretable decision traces~\cite{shi.etal_2026,lawrenceArgumentMiningSurvey2020}.

In real-world settings, AM relies on two important foundations: (i)~\textit{formal theoretical frameworks} that model argumentative interactions~\cite{gabbay.etal_2021,baroniHandbookFormalArgumentation2018}, and (ii)~\textit{data-driven methods} that leverage discourse cues to identify and classify arguments~\cite{budzynska.etal_2016,lawrenceArgumentMiningSurvey2020}. While substantial progress has been made in both formal argumentation~\cite{modgil.prakken_2014} and computational argument mining~\cite{lippi.torroni_2016,cabrioFiveYearsArgument2018}, effectively integrating these paradigms to produce interpretable, large-scale argument analysis remains an open problem~\cite{ruiz-dolz.etal_2022}.

Neuro-symbolic AI offers a promising direction for bridging formal and natural argumentation by combining the scalability of neural learning with the interpretability of symbolic reasoning~\cite{garcez.lamb_2023}. Such approaches have demonstrated potential for improving reliability and explainability across knowledge-intensive tasks~\cite{ulbricht.wallner_2022}. However, relatively little work has explored neuro-symbolic approaches for large-scale, real-world argument mining~\cite{ruiz-dolz.etal_2022,ulbricht.wallner_2022}, where systems must process raw, noisy discourse while producing formally grounded and traceable knowledge representations.

To address this gap, we propose a \textbf{Neuro-symbolic Argumentation Reasoning System} (NARS) consisting of two system components: a neural component that extracts bipolar argument structures from raw discussions using iterative LLM-based reasoning, and a symbolic component that applies gradual argumentation semantics to evaluate argumentative influence and derive structured knowledge. We evaluate NARS on the task of \textbf{persuasiveness detection}, where the objective is to determine whether the original user changes their viewpoint (henceforth referred to as the \textit{delta event}) during a discussion. Experiments are conducted on our novel CMV2BAS Dataset, derived from the Webis-CMV-20 corpus~\cite{alkhatib:2020b}, a large collection of moderated multi-threaded discussions from the ChangeMyView (CMV) subreddit.

To detect persuasiveness, our framework assigns each argumentative unit an \textit{a priori} strength representing its initial persuasive influence, which is then iteratively propagated through gradual semantics until equilibrium is reached. Since these \textit{a priori} strengths directly shape how influence evolves across the argument graph, we investigate multiple initialization strategies and analyze their impact on persuasiveness detection. Specifically, we address the following research questions:

\noindent\textbf{RQ1.} Can gradual argumentation semantics effectively model persuasiveness in long discussions?

\noindent\textbf{RQ2.} Does incorporating argument quality and structural centrality improve persuasion modeling under gradual semantics?

\noindent\textbf{RQ3.} Can large language models independently perform reliable persuasiveness reasoning, or are explicit symbolic reasoning frameworks necessary for modeling argumentative influence?

The main contributions of our work are:
\begin{enumerate}
    \item \textbf{CMV2BAS Dataset:} A dataset of long, multi-threaded CMV discussions annotated with LLM-derived bipolar argument structures and delta events.
    \item \textbf{NARS Framework:} An open-source, neuro-symbolic system for processing long discussions and determining persuasiveness.
    \item \textbf{Analytical Study:} A systematic evaluation of \textit{a priori strength} strategies and their impact on modeling argumentative influence and persuasiveness.
\end{enumerate}

\section{Background}

\subsection{Argumentation Semantics}
Formal argumentation theory offers several families of semantics for evaluating arguments, differing in how they model acceptability. \textbf{Extension-based semantics} identifies sets of jointly acceptable arguments --- known as extensions --- satisfying properties such as admissibility, groundedness, or stability~\cite{dung_1995,polberg2014extension}. While logic-driven, extensions yield binary accept/reject decisions and offer limited support for fine-grained argument comparison. \textbf{Ranking-based semantics} addresses this by inducing a total or partial ordering over arguments according to relative acceptability~\cite{10.1007/978-3-642-40381-1_11,heyninck.etal_2023}, enabling richer comparisons even in the presence of multiple extensions. However, rankings do not quantify absolute argumentative strength and disregard intrinsic argument properties, limiting their applicability in real-world settings where arguments vary in quality and certainty.

\textbf{Gradual semantics} addresses these limitations by assigning continuous acceptability values to arguments based on structural interactions via a gradual iterative process~\cite{amgoud.etal_2024,beuselinck.etal_2023}. This makes gradual semantics well-suited for natural argumentative discourse, where arguments are inherently qualitative and uncertainty is pervasive, especially in dynamic, non-monotonic settings~\cite{wang.shen_2024,potyka_2021}. Gradual semantics evaluates arguments through an iterative process that (i)~aggregates supporting and attacking influences, and (ii)~updates strength values until a stable equilibrium is reached~\cite{rago.etal_2025,baroni.etal_2019}.

A critical yet underexplored factor in applying gradual semantics to real-world discourse is the \textit{a priori} strength initialization of argument structures. Due to complexity of real-world discourse, prior work has been largely limited to controlled, synthetic settings with manually curated argument graphs~\cite{beuselinck.etal_2023,rago.etal_2025}, leaving open the question of how to effectively initialize and propagate strength in large-scale argument structures. In this work, we directly address this gap: we propose and evaluate multiple initialization strategies grounded in both semantic quality and structural centrality, and apply gradual semantics at scale to real-world online discussions. This sets our framework NARS among the first concrete steps toward operationalizing formal argumentation theory in practical, end-to-end argument analysis pipelines.

\subsection{Persuasiveness Detection}
Argumentation frameworks are well-suited for modeling non-monotonic reasoning, where accepted arguments may be revised as new evidence is introduced~\cite{amgoud.nouioua_2017}. This makes them naturally applicable to tasks such as belief revision~\cite{falappa.etal_2011}, polarity prediction~\cite{lenz.bergmann_2024}, and persuasiveness detection~\cite{verma.etal_2026,habernal.gurevych_2016}.

Persuasion is a central phenomenon in online discourse, particularly in domains such as politics, law, and debate~\cite{lukin.etal_2017}. Existing computational approaches typically model persuasiveness as a binary classification task using surface-level features or fine-tuned language models~\cite{tan.etal_2016,hidey.etal_2017,ji.etal_2018,bozdag.etal_2025}. However, persuasion in long-form discussions emerges gradually through multi-turn argumentative interactions rather than isolated discourse cues. Recent argumentation-aware approaches~\cite{ruiz-dolz.etal_2022} explore perusasion at macro-level, though remain limited in their application and annotated datasets. To address this our proposed system, NARS, models persuasiveness detection as a gradual shift in the argument strength of the central proposition (the main opinion to change) through argumentative influence propagation, where the human participant interactions act as indicators of supporting and attacking signals. For example, multiple attacks against a claim may collectively reduce its strength even when individual arguments are only moderately persuasive. Similarly, a strong argument asserts a strong influence on to its target. In this work, we focus on operationalizing existing formal gradual argumentation semantics over noisy real-world discourse for scalable macro-level persuasion analysis

\section{Data}

We source our data from the publicly available Webis ChangeMyView Corpus 2020 (Webis-CMV-20)~\cite{alkhatib:2020b}, which contains posts and comments from the ChangeMyView subreddit spanning 2005 to September 2017. The original corpus consists of 65,169 submissions and approximately 602,706 discussion threads replying directly to the original poster (OP). CMV is a heavily moderated subreddit with explicit discussion rules, where moderators actively remove any non-compliant comments.

For this study, we extract discussion threads satisfying several selection criteria to meet our experimental design. Each thread begins with an OP presenting a viewpoint intended to be challenged and consists of a tree-structured discussion containing replies and nested interactions. Threads are categorized into either \textbf{delta threads} (positive outcome), where the OP awards a explicit delta ($\Delta$) indicating a change in its viewpoint, or \textbf{non-delta threads} (negative outcome), where no delta is awarded. Given non-delta threads do not have clear endpoint, unlike delta threads, long threads are manually ended on the comment containing the 500th sentence. Additionally to narrow our search space, we apply topic filtering targeting discussions related to \textit{health}, \textit{migration}, \textit{climate}, and \textit{European integration}---topics that are politically polarizing and typically associated with higher argumentative activity. Finally, each thread must contain minimum five comments, spanning a minimum total of 30 sentences to ensure sufficient conversational activity.

All threads undergo preprocessing steps prior to analysis. Threads are pseudo-nymized to preserve user privacy, and no user profiling is performed in this study. Removed or deleted comments are retained to preserve reply structure integrity, although they are excluded from argument structure modeling. User mentions and quotations are replaced with the placeholder string ``previously stated...'' to maintain contextual relations between comments.

Following these steps, we identified 1,822 well-structured threads, of which 327 were delta threads. Due to the computational cost of our NARS pipeline, we sample 500 threads (250 delta- and 250 non-delta threads) for this study and release the broader data processing pipeline publicly.\footnote{\url{https://github.com/The-obsrvr/CMV-BAS-Data-Pipeline}}. Delta threads contain an average of 11.39 turns (maximum 106) and 78.12 sentences (maximum 508), while non-delta threads contain an average of 13.33 turns (maximum 106) and 107.90 sentences (maximum 608). These threads are subsequently processed through the proposed NARS framework to form our CMV2BAS.

\section{Methodology}
\label{sec:method}
Our proposed framework, NARS, consists of two primary components: a \textit{neural} component and a \textit{symbolic} component. The neural component employs a retrieval-augmented, iterative large language model (LLM) reasoning pipeline to extract coherent bipolar argument structures from multi-threaded discussions. These structures are subsequently processed by the symbolic component, which initializes each argumentative unit with an \textit{a priori} strength value representing its initial influence. Iterative influence propagation is then performed over the resulting argument graph using gradual argumentation semantics until a stable equilibrium state is reached.

\begin{figure*}[t]
    \centering
    \includegraphics[width=\textwidth]{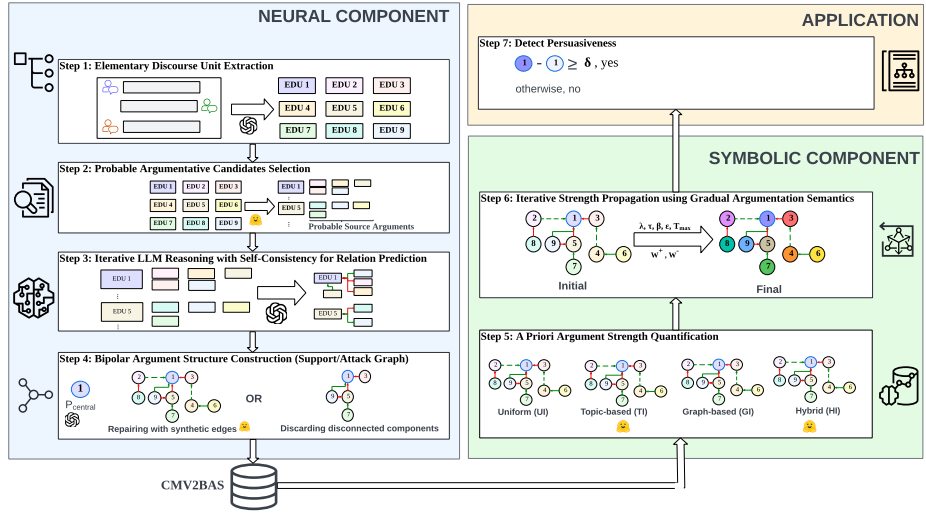}
    \caption{Overview of the NARS architecture, illustrating the neural component and its steps, the symbolic component and the downstream persuasiveness detection module.}
    \label{fig:sys}
\end{figure*}

The resulting strength dynamics are used to model persuasiveness as a gradual argumentative process rather than a single binary event. Figure~\ref{fig:sys} illustrates the overall NARS architecture, highlighting the interaction between the neural and symbolic components as well as the downstream persuasiveness detection task. Overall, the framework is organized into \textbf{seven} modular stages spanning argument extraction, relation prediction, graph construction, symbolic reasoning, and persuasiveness detection (see Table~\ref{tab:impl} for the complete system implementation details).

\textbf{Step 1. LLM-based EDU Extraction.} \textit{Elementary Discourse Units} are minimal discourse segments that express a single coherent proposition~\cite{saveleva.etal_2021}, typically split at clause boundaries and discourse connectives (e.g., \textit{and}, \textit{but}, \textit{or}).

We use a reasoning-based LLM to segment each discussion into EDUs. This approach improves robustness over rule-based splitting by handling implicit or non-syntactic boundaries common in social media text. For example, “Climate change is real and it threatens our future” is split into [“Climate change is real”, “and it threatens our future”]. If the LLM generation fails after two tries, a punctuation-based fallback logic is performed ensures full coverage.

\textbf{Step 2. Probable Argumentative Candidate Selection.}
Given the large size of individual discussion threads, open-source LLMs~($\leq30B$) often exhibit reasoning errors and reduced task alignment when dealing with long contexts~\cite{liu.etal_2024a}. To mitigate this, we adopt a retrieval-augmented strategy in which, for each EDU (the target), we select argumentatively relevant source candidates, referred to as Probable Argumentative Candidates (PACs) before the LLM reasoning step.

PACs are retrieved by $k$-nearest-neighbour search over sentence-transformer embeddings~\cite{reimers-2019-sentence-bert}, filtered by a minimum cosine similarity threshold, from which the top $n_{\text{semantic}}$ candidates are retained. To capture contextually relevant but semantically dissimilar EDUs (e.g., immediate neighbours: ["...I believe that is why I don't trust it"]), we additionally include up to $n_{\text{context}}$ subsequent EDUs not already selected. Finally, temporal directionality is enforced by restricting source candidates to EDUs occurring after the target EDU.

\textbf{Step 3. Iterative LLM Reasoning for Relation Prediction.} Given PACs for each EDU, an LLM (in \textit{thinking} mode) performs zero-shot classification to assign each candidate a \textit{support}, \textit{attack}, or \textit{neutral} (non - argumentative) relation toward its target EDU. Additionally in the prompts, the context surrounding the target EDU, up to two sentences before and after, is included to facilitate LLM reasoning.

To improve reliability under noisy long-context inputs, we use self-consistency scheme with two low-temperature inference passes for deterministic generation, followed by majority voting. If any, the disagreements are resolved via a third tie-breaking call, providing a form of weak supervision without task-specific training~\cite{du.etal_2024}. We retain only EDUs, henceforth becoming argumentative units, that participate in at least one support or attack relation, filtering out non-argumentative or isolated units.

\textbf{Step 4. Bipolar Argument Structure Construction.}
Predicted relations are consolidated into a bipolar argument graph $G(\mathcal{A}, \mathcal{E}, P_{\text{central}})$, where $\mathcal{A}$ denotes argumentative units, $\mathcal{E}$ support and attack relations, and $P_{\text{central}}$ the original poster's central proposition. $P_{\text{central}}$ is identified via a medium-thinking LLM prompt over opening-post units and manually validated; we further ensure it carries at least one incoming edge to enable influence propagation.

Since Step~3 may yield disconnected components, we consider two consolidation strategies: \textit{synthetic}, where disconnected components are linked to the main component via a single best similarity-based edge identified through cosine similarity over sentence-transformer embeddings, and \textit{discarded}, where only the component containing $P_{\text{central}}$ is retained. Graphs with fewer than three argumentative units are excluded, due to insufficient argumentative activity for meaningful evaluation.

\textbf{Step 5. A Priori Argument Strength Quantification.}
Each argumentative unit in $G$ carries some degree of persuasive influence within the discussion. To model how this influence affects the central proposition, arguments must first be assigned appropriate \textit{a priori}, strengths. Since arguments in online discussions are often implicit, incomplete, and vary in quality and structural importance, this initialization step critically captures these characteristics prior to influence propagation.

Formally, each argument graph is represented as an argumentation framework $AF = \langle \mathcal{A}, \mathcal{E}, \mathcal{I} \rangle$, where $\mathcal{A}$ denotes the set of arguments, $\mathcal{E}$ the set of support/attack relations, and $\mathcal{I}$ the initialization function assigning a priori argument strengths. We explore following complementary initialization strategies based on intrinsic and structural properties of the argument graph.

\begin{algorithm}[t]
\caption{A Priori Argument Strength Quantification}
\label{alg:strength_estimation}
\begin{algorithmic}[1]
\REQUIRE Units $\mathcal{A} = \{a_1, \dots, a_n\}$; relations $\mathcal{E}$; fine-tuned argument quality model; central proposition $P_{\text{central}}$
\ENSURE Normalized node strengths $S = \{s_1, \dots, s_n\}$

\STATE \textbf{UI scoring ($u_i'$):}
\FOR{each unit $a_i$}
    \STATE Set $u_i' = 1$
\ENDFOR
\STATE \textbf{TI scoring ($r_i'$):}
\FOR{each unit $a_i$}
    \STATE Compute pairwise quality score $r_i$ using fine-tuned argument quality model with input $(P_{\text{central}}, a_i)$
\ENDFOR
\STATE Normalize $r_i$ to obtain $r_i' \in [0,1]$
\STATE \textbf{GI scoring ($c_i'$):}
\STATE Construct directed graph $G = (\mathcal{A}, \mathcal{E})$
\STATE Compute PageRank centrality $c_i$ for each unit $a_i$
\STATE Normalize $c_i$ to obtain $c_i' \in [0,1]$
\STATE \textbf{HI scoring ($h_i'$):}
\STATE Compute undirected shortest path distance $d_i$ from each $a_i$ to $P_{\text{central}}$ in $G$
\STATE Set proximity weight $\alpha(d_i) = 1 - d_i / d_{\max}$
\STATE $h_i = r_i' + \alpha(d_i) \cdot c_i'$
\STATE Normalize $h_i$ to obtain $h_i' \in [0,1]$
\RETURN $S = \{(u_i', r_i', c_i', h_i')\}$
\end{algorithmic}
\end{algorithm}

\begin{enumerate}
    \item \textbf{Uniform Initialization (UI):}
    Following standard gradual semantics~\cite{potyka_2021}, each argumentative unit is assigned an identical prior strength of $1.0$ (see Algorithm~\ref{alg:strength_estimation}, lines 1--4), independent of its content or structure. This provides a neutral baseline where no prior bias is introduced on argumentative influence and serves as the baseline for \textbf{RQ1}.

    \item \textbf{Topic-based Initialization (TI):}
    A priori strengths are estimated using a DeBERTa-v3-small~\cite{he2021debertav3} pairwise argument-quality model fine-tuned on the Webis-ArgQuality-20 corpus~\cite{gienapp:2020}, which contains human-annotated relevance scores. This formulation explicitly scores each argumentative unit with respect to $P_{\text{central}}$, using topical relevance as a substitute for persuasive influence. While effective for explicit, well-formed arguments aligned with $P_{\text{central}}$, it is less reliable for implicit or context-dependent units (see Algorithm~\ref{alg:strength_estimation}, lines 5--9).

    \item \textbf{Graph-based Initialization (GI):}
    A priori strengths are computed using graph-based centrality measures, specifically PageRank in-degree centrality~\cite{page.etal_1999}, to capture the structural importance of the units within the argument graph. This strategy emphasizes arguments occupying influential structural positions irrespective of semantic quality, making it effective for identifying structurally central arguments but potentially weaker in semantic interpretation (see Algorithm~\ref{alg:strength_estimation}, lines 10--13).

    \item \textbf{Hybrid Initialization (HI):}
    Motivated by the idea that argumentative importance depends on both semantic relevance and graph structure, HI combines topic-based quality (TI) and structural relevance (GI) with proximity to $P_{\text{central}}$. It preserves semantic strength while incorporating a distance-weighted structural bias, prioritizing arguments that are both semantically relevant and centrally positioned in the graph as the most influential. Formally,

    \begin{equation}
    \mathrm{HI}(a) = \mathrm{normalise}\!\left(\mathrm{TI}(a) + \alpha(d)\cdot \mathrm{GI}(a)\right)
    \label{eq:distance_weighted_strength}
    \end{equation}
    where $\alpha(d) = 1-\frac{d}{d_{\max}}$ and $d$ is the shortest path distance to $P_{\mathrm{central}}$. This ensures that structural influence is strongest near $P_{\mathrm{central}}$ and diminishes for peripheral arguments, which typically have weaker structural impact (see Algorithm~\ref{alg:strength_estimation}, lines~14--18).
\end{enumerate}

Additionally, we consider a \textbf{Majority Vote (Ensemble)} strategy, which aggregates the binary persuasiveness predictions across all four initialization strategies via majority vote, predicting persuasiveness when at least two strategies vote yes.

Besides node initialization, relations are also initialized through weighted support and attack matrices, denoted by $W^{+}\in\mathbb{R}_{\geq 0}^{n\times n}$ and $W^{-}\in\mathbb{R}_{\geq 0}^{n\times n}$ respectively, encoding the degree to which arguments positively or negatively influence one another. Due to the substantial imbalance between support and attack relations in the extracted argument structures, attacking arguments may be overwhelmed by the support majority during propagation. To counter this, we initialize the attack matrix $W^{-}$ with a $3\times$ boosting factor, amplifying their influence in the propagation.

\textbf{Step 6. Iterative Strength Propagation using Gradual Semantics.}
\begin{algorithm}[t]
\caption{Iterative Strength Propagation using Gradual Semantics}
\label{alg:gradual_acceptability}
\begin{algorithmic}[1]
\REQUIRE Nodes $\mathcal{A}=\{a_1,\dots,a_n\}$; support weights $W^{+}\in\mathbb{R}_{\geq 0}^{n\times n}$; attack weights $W^{-}\in\mathbb{R}_{\geq 0}^{n\times n}$; prior strengths $b\in[0,1]^n$; damping $\lambda\in(0,1]$; temperature $\tau>0$; beta $\beta\geq0$; tolerance $\varepsilon>0$; maximum iterations $T_{\max}$
\ENSURE Strength scores $s^\ast \in [0,1]^n$; trajectory $\{s^{(t)}\}$

\STATE \textbf{Initialization:} $s^{(0)} \leftarrow b$; $t \leftarrow 0$
\REPEAT
    \STATE Compute aggregated support and attack:
    \[
      p^{(t)} = (W^+)^\top s^{(t)}, \quad q^{(t)} = (W^-)^\top s^{(t)}
    \]
    \STATE Net influence: $u^{(t)} = \beta\, b + p^{(t)} - q^{(t)}$
    \STATE Gradual function (sigmoid): $\sigma_\tau(x) = \frac{1}{1 + e^{-x/\tau}}$
    \STATE Proposed update: $\tilde{s}^{(t+1)} = \sigma_\tau\!\big(u^{(t)}\big)$
    \STATE Damped iterate: $s^{(t+1)} = (1-\lambda)\, s^{(t)} + \lambda\, \tilde{s}^{(t+1)}$
    \STATE $t \leftarrow t+1$
\UNTIL{$\|s^{(t)} - s^{(t-1)}\|_\infty < \varepsilon$ \textbf{ or } $t \ge T_{\max}$}
\STATE \textbf{Return} $s^\ast \leftarrow s^{(t)}$
\end{algorithmic}
\end{algorithm}

We apply gradual semantics~\cite{amgoud.ben-naim_2018a,baroni.etal_2019} (Algorithm~\ref{alg:gradual_acceptability}) to compute posterior (final) argument strengths. At each iteration, the framework \textit{aggregates} the supporting and attacking influences via the relation weights, \textit{propagates} them across the graph, and \textit{updates} argument strengths through a sigmoid-based gradual function~\cite{potyka_2021}. This repeats until strength values converge to a stable equilibrium, enabling a macro-level, context-sensitive evaluation of persuasive influence across the full argument structure.

The algorithm is governed by five hyperparameters: $\lambda$ controls how much newly propagated information updates current strengths (line 6); $\beta$ retains a fraction of the a priori strength throughout propagation (line 4); $\tau$ governs the sharpness of the gradual update function (line 5); and $\varepsilon$, $T_{\max}$ define the convergence criterion (line 7). Values have been taken from existing semantics and tuned empirically (see Table~\ref{tab:impl}) and we leave a more detailed investigation of their effects for future work.

\textbf{Step 7. Persuasiveness Detection.} Persuasiveness is modeled as a \textit{significant posterior reduction in the graded strength value of the central proposition within the given argumentation framework}. Let $b_{central}$ and $s^{*}_{central}$ denote the prior and posterior strengths of $P_\text{central}$ after the iterative symbolic reasoning. A discussion is considered persuaded if the change in strength exceeds a predefined \textbf{persuasiveness sensitivity} threshold, denoted by $\delta$:

\[
\text{Persuaded},\left|\Delta\!\left(P_{\mathrm{central}}\right)\right|
=
\begin{cases}
yes, & \text{if }
\left|s_{\mathrm{central}}^{*}-b_{\mathrm{central}}\right|
\geq \delta \\[4pt]
no, & \text{otherwise}
\end{cases}
\]

Larger values of $\delta~(\geq0.5)$ impose stricter, more conservative criteria for persuasiveness, whereas smaller values may allow even subtle shifts in $P_\text{central}$'s argumentative strength to be considered sufficiently persuasive. In this work, we analyze the effect of varying $\delta$ on persuasiveness detection.

\section{Experimental Setup}
\label{sec:exp}

This study primarily focuses on Step~5 of our framework, where we investigate how \textit{a priori} argument strength can be modeled to capture the persuasive influence of argumentative units and subsequently propagated through the argument structure to infer persuasiveness. We implement the four initialization strategies introduced in Step~5 of Section~\ref{sec:method} and evaluate their effectiveness in addressing \textbf{RQ1} and \textbf{RQ2}.

To address \textbf{RQ3}, we further investigate whether modern LLMs can independently perform persuasiveness detection without explicit symbolic reasoning mechanisms. Specifically, we experiment with two state-of-the-art, open-source thinking LLMs ($\leq30B$, due to computational limit), \textit{Qwen3.6:27B}~\cite{qwen36_27b} and \textit{GPT-oss:20B}~\cite{openai2025gptoss120bgptoss20bmodel} (medium thinking), under three zero-shot prompting settings: (i) using only the discussion text, (ii) using the discussion together with the bipolar argument structure (synthetic version) generated in Step~4, and (iii) additionally incorporating the \textit{a priori} argumentative strength values (HI) produced in Step~5. Although prompts are adapted to each setting, the models are constrained to output only a binary persuasiveness decision.

Given the lack of existing work on this data and task, we benchmark our framework against two naive baselines: a random classifier and a feature-based logistic regression model using surface-level discussion properties. The logistic regression baseline uses four features---direct attacks on $P_{\text{central}}$, total attacks, support--attack ratio, and the in-degree of $P_{\text{central}}$---to estimate persuasiveness. To train this baseline, we additionally processed 100 discussions from the original corpus.

\begin{table}[t]
\centering
\small
\begin{tabularx}{\columnwidth}{lX}
\toprule
\textbf{Step} & \textbf{Configuration} \\
\midrule
\textbf{S1} EDU Extraction
& Qwen3:27B~\cite{qwen36_27b}; one-shot, zero-thinking; temp=0.1; batch=200 sentences; max 8,192 tokens; punctuation fallback \\
\midrule
\textbf{S2} PAC Selection
& \texttt{all-MiniLM-L6-v2}~\cite{reimers-2019-sentence-bert}; $k$=25; sim. threshold=0.45; $n_\text{semantic}$=10, $n_\text{context}$=5 \\
\midrule
\textbf{S3} Relation Prediction
& GPT-oss:20B (medium thinking)~\cite{openai2025gptoss120bgptoss20bmodel}; batch=10 EDUs $\pm$2 sentences for context; self-consistency with temp=\{0.1, 0.2, 0.15 (tie-breaker)\}; max 8,192 tokens; JSON structured output \\
\midrule
\textbf{S4} BAS Construction
& $P_\text{central}$ via GPT-oss:20B (medium thinking); \textit{discarded}: only main component containing $P_\text{central}$; \textit{synthetic}: cosine-similarity repair via \texttt{all-MiniLM-L6-v2}; min. 3 nodes in final graph \\
\midrule
\textbf{S5} Strength Init.
& UI / GI / TI / HI / Ensemble; $W^+$=1.0/$W^-$=3.0; synthetic edge weight=0.7 \\
\midrule
\textbf{S6} Gradual Semantics
& $\lambda$=0.5, $\tau$=0.2, $\beta$=0.1, $\varepsilon$=1e-6, $T_{\max}$=1,000 \\
\midrule
\textbf{S7} Persuasiveness
& $\delta \in \{0.1, 0.3, 0.5, 0.7, 0.8\}$ \\
\bottomrule
\end{tabularx}
\caption{Implementation details per pipeline step, reflecting the most efficient configuration identified through empirical investigation.}
\label{tab:impl}
\end{table}

All experiments are conducted on a 48GB NVIDIA Ampere A40 GPU within a Docker-based environment. For the steps involving LLMs, the models and their inference are performed through the Ollama client~\cite{ollama2025}. Table~\ref{tab:impl} summarizes our final, most-efficient system configuration for each step. Full implementation details and source code are available in our system repository (see Footnote~\ref{fn:repo}).

\subsection{CMV2BAS Dataset}

We construct and release CMV2BAS as a large-scale benchmark for persuasive argumentation analysis and neuro-symbolic reasoning. Starting from 500 raw discussion threads, we first segmented the text into 45,689 Elementary Discourse Units (EDUs), enabling fine-grained modeling of long and noisy social media discourse. Step~2 retrieved 350,101 probable argumentative candidate pairs (PACs) based on the semantic and contextual similarity. Step~3 performs iterative LLM-based relation prediction with self-consistent reasoning to classify support and attack relations, forming the computational bottleneck (9--12 minutes per ~10 EDUs). Step~4 consolidates these predictions into connected bipolar argument structures centered on the central proposition, filtering out the smaller~($\leq3$ units) discussion threads.

This pipeline yields 384 valid discussions from 500 CMV threads (202 delta-event, 182 non-delta-event). The final dataset contains 24,407 argumentative units and 48,166 support/attack relations. Both synthetic and discarded graph variants are retained in the dataset. Each discussion is enriched with metadata including the central proposition, speaker participation statistics, conversational structure, symbolic reasoning outputs, and persuasiveness labels. Outputs were sampled after each step for manual validation and found to be largely stable, with minor inconsistencies handled in post-processing.

In summary, CMV2BAS is a large-scale benchmark of long-form persuasive discussions with structured argument graphs and symbolic reasoning traces, supporting evaluation of argument structure modeling and persuasiveness detection in complex discourse.

\section{Results}

We evaluate different modeling paradigms on persuasiveness detection, formulated as a binary classification task, using Accuracy, F1-score, and Area Under the ROC Curve (AUC). Accuracy measures the overall correctness, F1-score captures the precision--recall trade-off, and AUC evaluates the model's discriminative power between persuasive and non-persuasive discussions. Since the symbolic propagation mechanism is deterministic given a fixed argument graph and initialization strategy, the framework does not exhibit variance across random seeds. Instead, to assess robustness under dataset variability, we report bootstrap confidence intervals ($B{=}1000$)~\cite{https://doi.org/10.1111/j.1740-9713.2004.00067.x} over evaluation metrics.

Beyond classification performance, we analyze the effect of the persuasiveness sensitivity $\delta$ and study how NARS behaves under varying $\delta$. We additionally perform qualitative analysis on the generated argument structures and symbolic reasoning traces to identify common failure modes, including dominant support relations and signal attenuation across long argumentative paths.

\subsection{Persuasiveness Detection}

Table~\ref{tab:persuasion_results} reports the performance of different modeling paradigms on persuasiveness detection.

\begin{table}[t]
\centering
\small
\begin{tabularx}{\columnwidth}{llccc}
\toprule
\textbf{Category} & \textbf{Method} & \textbf{Accuracy} & \textbf{F1} & \textbf{AUC} \\
\midrule
\multirow{2}{*}{Naive Baseline}
& Random                          & 0.5025 & 0.5266 & 0.5000 \\
& Log. Regression  & 0.5392 & 0.5635 & 0.5869 \\
\midrule
\multirow{5}{*}{NARS (discarded); $\delta=0.3$}
& UI         & 0.5417$_{0.489}^{0.591}$ & 0.5165$_{0.451}^{0.575}$ & 0.5620$_{0.502}^{0.616}$ \\
& GI           & 0.5000$_{0.451}^{0.549}$ & 0.5450$_{0.486}^{0.604}$ & 0.4697$_{0.413}^{0.529}$ \\
& TI           & 0.4635$_{0.411}^{0.513}$ & 0.4976$_{0.434}^{0.556}$ & 0.4638$_{0.409}^{0.525}$ \\
& HI          & 0.4505$_{0.401}^{0.505}$ & 0.4548$_{0.392}^{0.517}$ & 0.4605$_{0.404}^{0.516}$ \\
& Ensemble            & 0.4922$_{0.443}^{0.542}$ & 0.5412$_{0.482}^{0.601}$ & 0.4757$_{0.421}^{0.533}$ \\
\midrule
\multirow{5}{*}{NARS (synthetic); $\delta=0.3$}
& UI         & \textbf{0.5521$_{0.502}^{0.602}$} & 0.5196$_{0.455}^{0.580}$ & \textbf{0.5643$_{0.503}^{0.618}$} \\
& GI           & 0.5000$_{0.453}^{0.549}$ & 0.5493$_{0.491}^{0.608}$ & 0.4691$_{0.411}^{0.530}$ \\
& TI           & 0.4609$_{0.411}^{0.513}$ & 0.5106$_{0.446}^{0.568}$ & 0.4701$_{0.411}^{0.526}$ \\
& HI          & 0.4896$_{0.440}^{0.539}$ & 0.5051$_{0.440}^{0.562}$ & 0.5108$_{0.454}^{0.566}$ \\
& Ensemble            & 0.4974$_{0.448}^{0.549}$ & \textbf{0.5682$_{0.510}^{0.621}$} & 0.5135$_{0.458}^{0.573}$ \\
\midrule
\multirow{3}{*}{Qwen3.6:27B (0-shot)}
& Discussion                      & 0.4914 & 0.2525 & 0.5227 \\
& + structure (synth)         & 0.4845 & 0.2647 & 0.5140 \\
& + HI prior strength          & 0.4880 & 0.2660 & 0.5178 \\
\midrule
\multirow{3}{*}{GPT-oss:20B (0-shot)}
& Discussion                      & 0.4674 & 0.1341 & 0.5041 \\
& + structure (synth)         & 0.4742 & 0.1639 & 0.5099 \\
& + HI prior strength           & 0.4708 & 0.1250 & 0.5085 \\
\bottomrule
\end{tabularx}
\caption{Evaluation of persuasiveness detection across modeling paradigms on CMV2BAS ($n=384$). NARS results report mean with 95\% confidence interval through bootstrapping ($_{lower}^{upper}$, $B{=}1000$) at $\delta=0.3$. LLM baselines report point estimates only. \textbf{Bold} denotes best per column among NARS configurations.}
\label{tab:persuasion_results}
\end{table}

\textbf{RQ1 — Gradual argumentation semantics for persuasiveness.}
NARS with Uniform Initialization (UI) shows that gradual argumentation semantics can capture persuasiveness signals beyond naive baselines. At $\delta=0.3$, UI achieves F1=0.5196 $[0.455, 0.580]$ and AUC=0.5643 $[0.503, 0.618]$ under the synthetic variant, with the AUC interval lying above random performance (0.500). The wide confidence intervals indicate a modest signal, and results are broadly comparable to the logistic regression baseline, suggesting that standard gradual semantics captures a weak but non-trivial persuasiveness signal.

\textbf{RQ2 — Role of argument quality and structural centrality.}
Initialization strategies have a limited and inconsistent effect on performance. UI achieves the best AUC despite its simplicity, and overlapping confidence intervals across all strategies  strong conclusions. GI yields a higher F1 than UI (0.5493 vs.\ 0.5196), while TI and HI do not consistently improve over simpler alternatives. The Ensemble achieves the highest F1 overall (0.5682), though its interval overlaps with other strategies. Synthetic graph variants show marginal gains over discarded variants under UI and GI, being more structurally oriented, with negligible differences in other configurations.

\textbf{RQ3 — LLM-based vs.\ symbolic reasoning.}
Both LLMs, with thinking mode, perform well below the random baseline in F1 (Qwen: 0.2660, GPT-oss: 0.1639 at best), with AUC near 0.50. Augmenting prompts with argument structure and strength scores yields only marginal gains, suggesting that zero-shot LLMs struggle to exploit structured argumentation representations for this task.

The large bootstrap confidence intervals (typically $\pm$0.05--0.06) reflect the difficulty of the task and the moderate corpus size. Overlapping intervals across NARS configurations indicate that differences between individual strategies are not statistically pronounced, with structural features appearing to be the dominant factor for the initialization choice.

\subsection{Persuasiveness Sensitivity}

\begin{figure}[t]
    \centering
    \begin{subfigure}[b]{0.48\columnwidth}
        \centering
        \includegraphics[width=\textwidth]{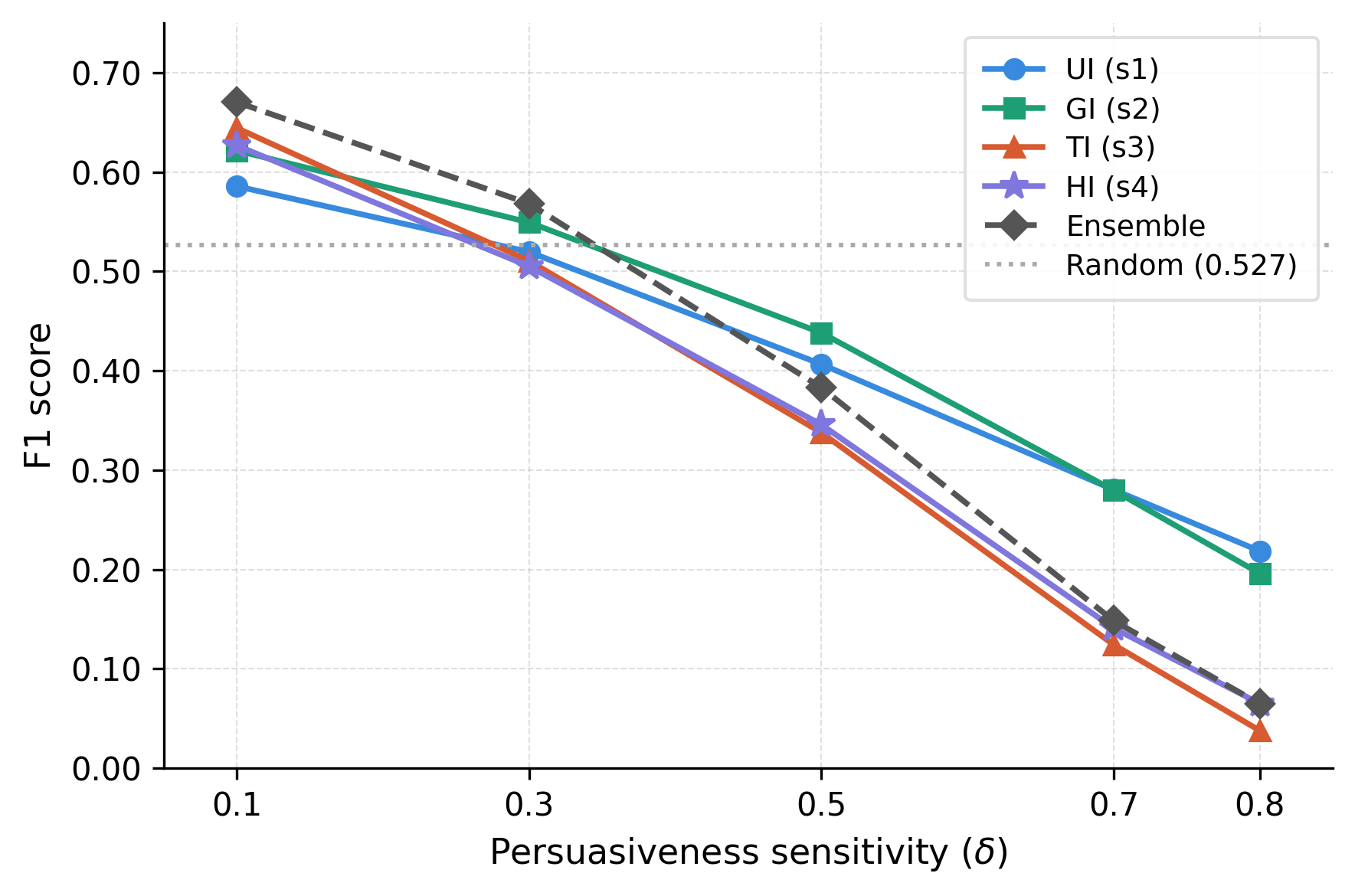}
        \caption{Relationship between $\delta$ and F1-score.}
        \label{fig:delta_f1}
    \end{subfigure}
    \hfill
    \begin{subfigure}[b]{0.48\columnwidth}
        \centering
        \includegraphics[width=\textwidth]{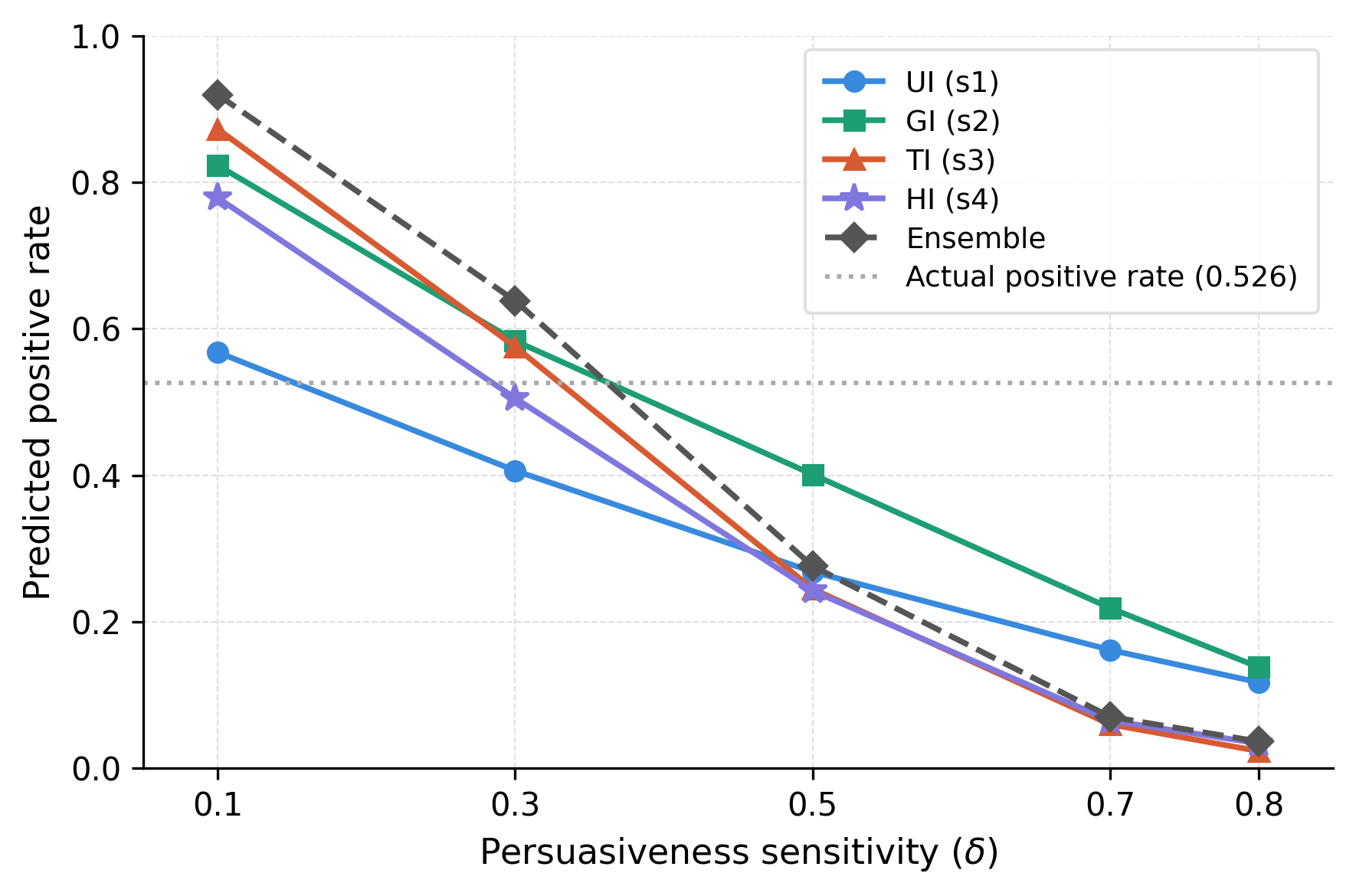}
        \caption{Predicted positive rate vs.\ $\delta$.}
        \label{fig:delta_bias}
    \end{subfigure}
    \caption{Effect of persuasiveness sensitivity threshold $\delta$ on F1 score (left) and prediction bias (right) across initialization strategies in synthetic mode ($n=384$).}
    \label{fig:sensitivity_analysis}
\end{figure}

To evaluate the robustness of NARS under varying persuasiveness sensitivity thresholds, we analyse the relationship between $\delta$, F1-score, and the predicted positive rate -- the rate at which the system classifies a discussion as persuasive. Figure~\ref{fig:sensitivity_analysis} presents the results.

At lower values of $\delta$, all initialization strategies exhibit a substantially inflated tendency to predict the positive class. As shown in Figure~\ref{fig:delta_bias}, most strategies intersect the true positive rate (0.526) between $\delta=0.3$ and $\delta=0.5$, indicating that this range constitutes the most calibrated operating region for persuasiveness detection. As $\delta$ increases, the framework becomes progressively conservative, requiring stronger accumulated argumentative influence to classify a discussion as persuasive, and F1 degrades accordingly as the system defaults toward the negative class.

Among the initialization strategies, UI followed by GI demonstrate the most stable F1 and AUC across the sensitivity range. In contrast, TI, HI, and the Ensemble exhibit a sharp performance decline beyond $\delta=0.5$, suggesting that semantic and hybrid initializations propagate insufficient influence to maintain discriminability at stricter thresholds. Overall, $\delta$ acts as a tunable hyperparameter governing the precision--recall trade-off, with $\delta=0.3$ offering the most balanced configuration among our experiments.

\paragraph{Error Analysis.}
Through qualitative analysis, we identify two primary failure modes that limit effective attack signal propagation. \textit{Support dominance}: in highly support-dominant discussions, the overwhelming volume of supporting arguments suppresses any influence of attacking arguments on $P_{\text{central}}$, regardless of their argumentative quality. To partially mitigate this effect, we experimentally applied a $3\times$ weight multiplier to attack edges; however, the resulting performance gains remained marginal, suggesting that the opposing signals were insufficient against the large volume of supporting signals to meaningfully influence the final propagation dynamics.

\textit{Signal attenuation}: in long argumentative chains, the influence of peripheral attack nodes progressively weakens before reaching $P_{\text{central}}$, reducing their contribution to the final acceptability score. Both failure modes highlight a broader open challenge in modeling argumentative attacks within real-world discourse.

\begin{figure}[t]
    \centering
    \begin{subfigure}[b]{0.48\columnwidth}
        \centering
        \includegraphics[width=\textwidth]{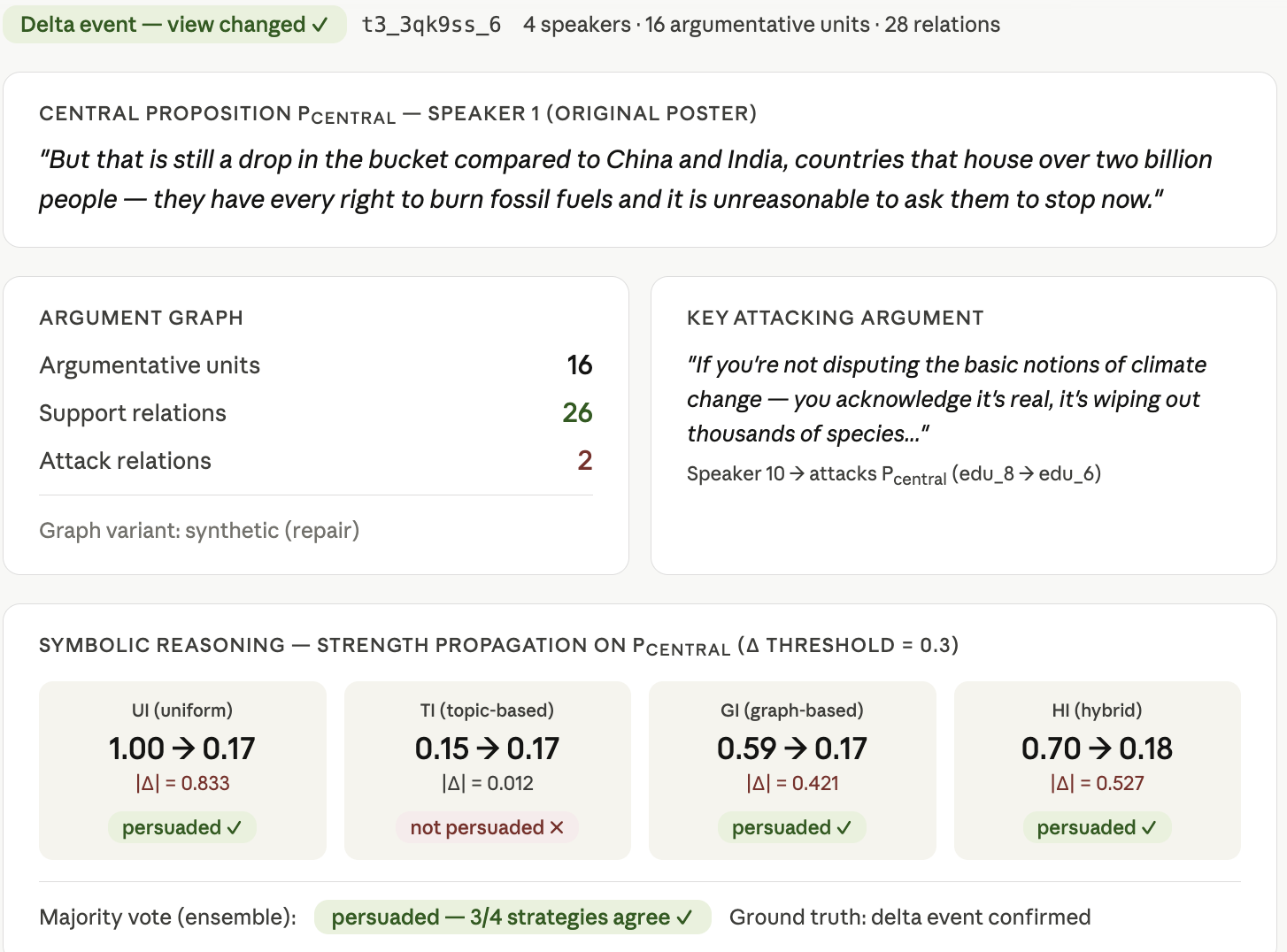}
        \caption{Correct Outcome}
        \label{fig:correct}
    \end{subfigure}
    \hfill
    \begin{subfigure}[b]{0.48\columnwidth}
        \centering
        \includegraphics[width=\textwidth]{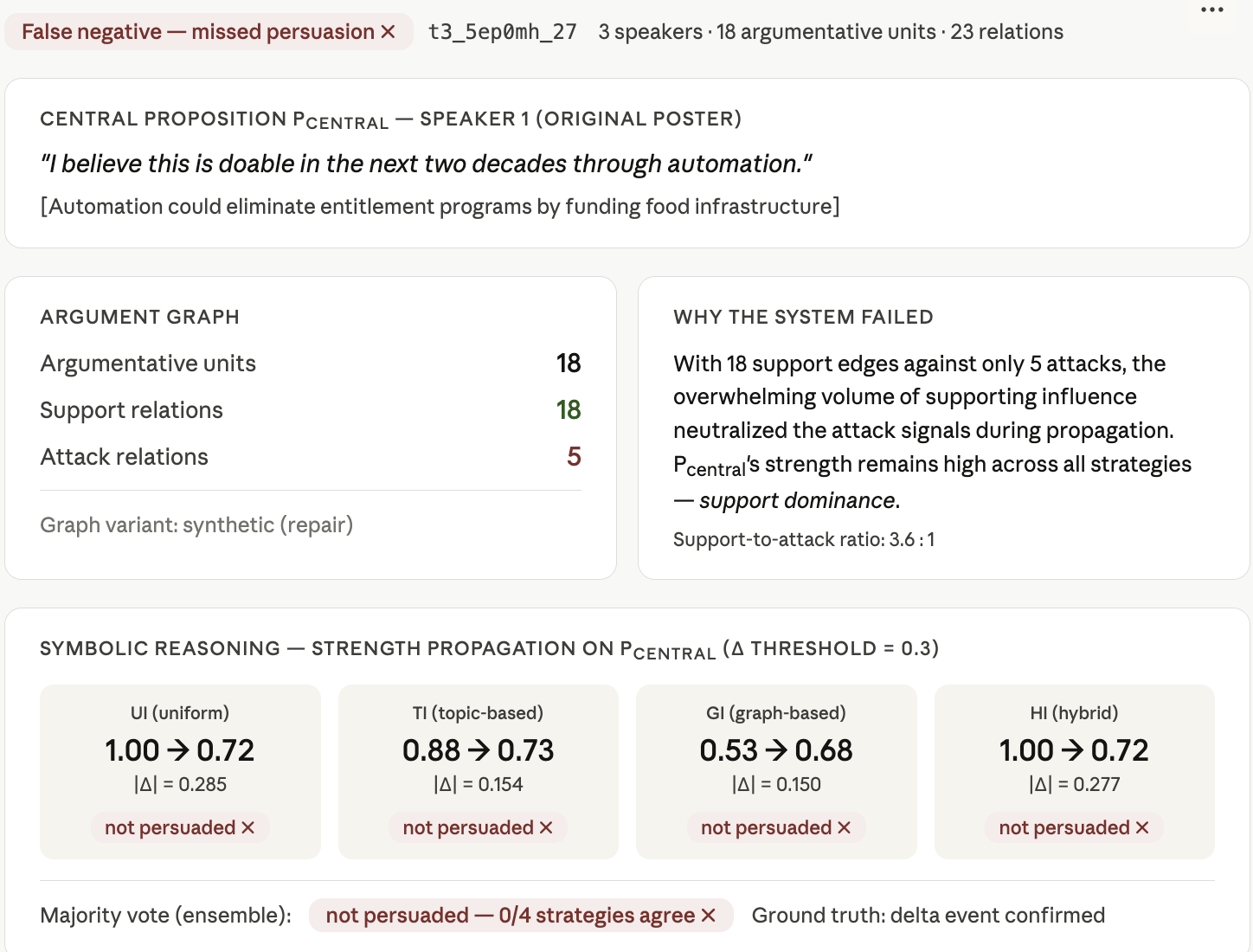}
        \caption{Failed Outcome.}
        \label{fig:failed}
    \end{subfigure}
    \caption{NARS analysis on two discussion threads}
    \label{fig:sample}
\end{figure}

\section{Conclusion}

We presented NARS, a neuro-symbolic system for argument structure extraction and macro-level persuasiveness detection over long, multi-threaded online discussions. We introduced CMV2BAS, a dataset of 384 discussions annotated with LLM-derived bipolar argument structures and delta events, which we release publicly alongside the framework. Figure~\ref{fig:sample} presents an infographic illustrating the NARS analysis of discussion threads taken from CMV2BAS. In summary, NARS achieves F1=0.57 and AUC=0.56 at $\delta=0.3$, moderately outperforming both feature-based and random baselines, while zero-shot LLMs fail to surpass random performance even when augmented with extracted argument structures and strength scores. Although NARS remains limited in practical deployment, our results demonstrate the feasibility of applying formal argumentation semantics to noisy real-world discourse for scalable macro-level persuasion analysis.

Our study underscores that strength initialization remains the primary bottleneck of the neuro-symbolic framework. Quantifying the influence and relative strength of arguments is a challenging problem, particularly when structural and quality-based initialization may introduce noisy or conflicting propagation patterns. In addition, the current NARS framework exhibits performance degradation over large argumentative distances, leading to signal attenuation across long argument chains.

Future work should explore recursive and hierarchical reasoning and more structure-oriented initialization strategies in which smaller argumentative substructures are processed independently before aggregation, helping preserve local argumentative activity and influence signals in increasingly large and complex discussions.

\begin{credits}
\subsubsection{\ackname}
This research work has received funding from the European Union's Horizon Europe research and innovation programme under the Marie Skłodowska-Curie Grant Agreement No. 101073351.

\end{credits}
%
%
%
\bibliographystyle{splncs04}
\bibliography{custom}

@inproceedings{heyninck.etal_2023,
  title = {Ranking-Based {{Argumentation Semantics Applied}} to {{Logical Argumentation}}},
  booktitle = {Proceedings of the {{Thirty-Second International Joint Conference}} on {{Artificial Intelligence}}},
  author = {Heyninck, Jesse and Raddaoui, Badran and Stra{\ss}er, Christian},
  year = 2023,
  month = aug,
  pages = {3268--3276},
  publisher = {International Joint Conferences on Artificial Intelligence Organization},
  address = {Macau, SAR China},
  doi = {10.24963/ijcai.2023/364},
  urldate = {2025-09-09},
  isbn = {978-1-956792-03-4},
  langid = {english}
}

@inproceedings{saveleva.etal_2021,
  title = {Discourse-Based {{Argument Segmentation}} and {{Annotation}}},
  booktitle = {Proceedings of the 17th {{Joint ACL}} - {{ISO Workshop}} on {{Interoperable Semantic Annotation}}},
  author = {Saveleva, Ekaterina and Petukhova, Volha and Mosbach, Marius and Klakow, Dietrich},
  year = 2021,
  pages = {41--53},
  langid = {english}
}

@InProceedings{alkhatib:2020b,
  author =                   {Khalid Al-Khatib and Michael V{\"o}lske and Shahbaz Syed and Nikolay Kolyada and Benno Stein},
  booktitle =                {58th Annual Meeting of the Association for Computational Linguistics (ACL 2020)},
  ids =                      {stein:2020l},
  month =                    jul,
  pages =                    {7067--7072},
  publisher =                {Association for Computational Linguistics},
  site =                     {Seattle, USA},
  title =                    {{Exploiting Personal Characteristics of Debaters for Predicting Persuasiveness}},
  url =                      {https://aclanthology.org/2020.acl-main.632/},
  year =                     2020
}

@inproceedings{du.etal_2024,
  title = {Improving Factuality and Reasoning in Language Models through Multiagent Debate},
  booktitle = {Proceedings of the 41st {{International Conference}} on {{Machine Learning}}},
  author = {Du, Yilun and Li, Shuang and Torralba, Antonio and Tenenbaum, Joshua B. and Mordatch, Igor},
  year = 2024,
  month = jul,
  series = {{{ICML}}'24},
  volume = {235},
  pages = {11733--11763},
  publisher = {JMLR.org},
  address = {Vienna, Austria},
  urldate = {2026-05-20}
}

@article{beuselinck.etal_2023,
  title = {A {{Principle-based Account}} of {{Self-attacking Arguments}} in {{Gradual Semantics}}},
  author = {Beuselinck, Vivien and Delobelle, J{\'e}r{\^o}me and Vesic, Srdjan},
  year = 2023,
  month = mar,
  journal = {Journal of Logic and Computation},
  volume = {33},
  number = {2},
  pages = {230--256},
  issn = {0955-792X, 1465-363X},
  doi = {10.1093/logcom/exac093},
  urldate = {2025-11-24},
  copyright = {https://academic.oup.com/journals/pages/open\_access/funder\_policies/chorus/standard\_publication\_model},
  langid = {english}
}

@article{wang.shen_2024,
  title = {Bilateral {{Gradual Semantics}} for {{Weighted Argumentation}}},
  author = {Wang, Zongshun and Shen, Yuping},
  year = 2024,
  month = mar,
  journal = {Proceedings of the AAAI Conference on Artificial Intelligence},
  volume = {38},
  number = {9},
  pages = {10732--10739},
  issn = {2374-3468},
  doi = {10.1609/aaai.v38i9.28945},
  urldate = {2025-11-05},
  copyright = {Copyright (c) 2024 Association for the Advancement of Artificial Intelligence},
  langid = {english}
}

@article{potyka_2021,
  title = {Interpreting {{Neural Networks}} as {{Quantitative Argumentation Frameworks}}},
  author = {Potyka, Nico},
  year = 2021,
  month = may,
  journal = {Proceedings of the AAAI Conference on Artificial Intelligence},
  volume = {35},
  number = {7},
  pages = {6463--6470},
  issn = {2374-3468, 2159-5399},
  doi = {10.1609/aaai.v35i7.16801},
  urldate = {2025-11-05},
  langid = {english}
}

@inproceedings{amgoud.etal_2024,
  title = {Higher-{{Order Argumentation Frameworks}}: {{Principles}} and {{Gradual Semantics}}},
  shorttitle = {Higher-{{Order Argumentation Frameworks}}},
  booktitle = {Thirty-{{Third International Joint Conference}} on {{Artificial Intelligence}}},
  author = {Amgoud, Leila and Doder, Dragan and {Lagasquie-Schiex}, Marie-Christine},
  year = 2024,
  month = aug,
  volume = {4},
  pages = {3224--3231},
  issn = {1045-0823},
  doi = {10.24963/ijcai.2024/357},
  urldate = {2026-05-21},
  langid = {english}
}

@article{amgoud.nouioua_2017,
  title = {An Argumentation System for Defeasible Reasoning},
  author = {Amgoud, Leila and Nouioua, Farid},
  year = 2017,
  month = jun,
  journal = {International Journal of Approximate Reasoning},
  volume = {85},
  pages = {1--20},
  issn = {0888-613X},
  doi = {10.1016/j.ijar.2017.02.009},
  urldate = {2024-04-17}
}

@inproceedings{10.1007/978-3-642-40381-1_11,
  title = {Ranking-Based Semantics for Argumentation Frameworks},
  booktitle = {Scalable Uncertainty Management},
  author = {Amgoud, Leila and {Ben-Naim}, Jonathan},
  editor = {Liu, Weiru and Subrahmanian, V. S. and Wijsen, Jef},
  year = 2013,
  pages = {134--147},
  publisher = {Springer Berlin Heidelberg},
  address = {Berlin, Heidelberg},
  isbn = {978-3-642-40381-1}
}

@incollection{polberg2014extension,
  title={Extension--based semantics of abstract dialectical frameworks},
  author={Polberg, Sylwia},
  booktitle={STAIRS 2014},
  pages={240--249},
  year={2014},
  publisher={IOS Press}
}

@misc{rago.etal_2025,
  title = {A {{Methodology}} for {{Incompleteness-Tolerant}} and {{Modular Gradual Semantics}} for {{Argumentative Statement Graphs}}},
  author = {Rago, Antonio and Vasileiou, Stylianos Loukas and Toni, Francesca and Son, Tran Cao and Yeoh, William},
  year = 2025,
  month = aug,
  number = {arXiv:2410.22209},
  eprint = {2410.22209},
  primaryclass = {cs},
  publisher = {arXiv},
  doi = {10.48550/arXiv.2410.22209},
  urldate = {2025-11-24},
  archiveprefix = {arXiv},
  langid = {english}
}

@article{lawrenceArgumentMiningSurvey2020,
  ids = {lawrence.reed_2020,lawrence2020argument},
  title = {Argument {{Mining}}: {{A Survey}}},
  shorttitle = {Argument {{Mining}}},
  author = {Lawrence, John and Reed, Chris},
  year = 2020,
  month = jan,
  journal = {Computational Linguistics},
  volume = {45},
  number = {4},
  pages = {765--818},
  issn = {0891-2017, 1530-9312},
  doi = {10.1162/coli_a_00364},
  urldate = {2024-01-17},
  langid = {english}
}

@article{ulbricht.wallner_2022,
  title = {Strongly Accepting Subframeworks: {{Connecting}} Abstract and Structured Argumentation},
  shorttitle = {Strongly Accepting Subframeworks},
  author = {Ulbricht, M. and Wallner, J.P.},
  year = 2022,
  journal = {Frontiers in Artificial Intelligence and Applications},
  volume = {353},
  pages = {320--331},
  issn = {0922-6389},
  doi = {10.3233/FAIA220163},
  isbn = {9781643683065},
  langid = {english}
}

@misc{qwen36_27b,
  title = {{Qwen3.6-27B}: Flagship-Level Coding in a 27B Dense Model},
  url = {https://qwen.ai/blog?id=qwen3.6-27b},
  author = {{Qwen Team}},
  month = {April},
  year = {2026}
}

@inproceedings{hidey.etal_2017,
  title = {Analyzing the {{Semantic Types}} of {{Claims}} and {{Premises}} in an {{Online Persuasive Forum}}},
  booktitle = {Proceedings of the 4th {{Workshop}} on {{Argument Mining}}},
  author = {Hidey, Christopher and Musi, Elena and Hwang, Alyssa and Muresan, Smaranda and McKeown, Kathy},
  year = 2017,
  pages = {11--21},
  publisher = {Association for Computational Linguistics},
  address = {Copenhagen, Denmark},
  doi = {10.18653/v1/W17-5102},
  urldate = {2024-01-17},
  langid = {english}
}

@misc{bozdag.etal_2025,
  title = {Must {{Read}}: {{A Systematic Survey}} of {{Computational Persuasion}}},
  shorttitle = {Must {{Read}}},
  author = {Bozdag, Nimet Beyza and Mehri, Shuhaib and Yang, Xiaocheng and Ha, Hyeonjeong and Cheng, Zirui and Durmus, Esin and You, Jiaxuan and Ji, Heng and Tur, Gokhan and {Hakkani-T{\"u}r}, Dilek},
  year = 2025,
  month = may,
  number = {arXiv:2505.07775},
  eprint = {2505.07775},
  primaryclass = {cs},
  publisher = {arXiv},
  doi = {10.48550/arXiv.2505.07775},
  urldate = {2026-03-03},
  archiveprefix = {arXiv}
}

@inproceedings{ji.etal_2018,
  title = {Incorporating {{Argument-Level Interactions}} for {{Persuasion Comments Evaluation}} Using {{Co-attention Model}}},
  booktitle = {Proceedings of the 27th {{International Conference}} on {{Computational Linguistics}}},
  author = {Ji, Lu and Wei, Zhongyu and Hu, Xiangkun and Liu, Yang and Zhang, Qi and Huang, Xuanjing},
  editor = {Bender, Emily M. and Derczynski, Leon and Isabelle, Pierre},
  year = 2018,
  month = aug,
  pages = {3703--3714},
  publisher = {Association for Computational Linguistics},
  address = {Santa Fe, New Mexico, USA},
  urldate = {2026-05-21}
}

@inproceedings{lukin.etal_2017,
  title = {Argument {{Strength}} Is in the {{Eye}} of the {{Beholder}}: {{Audience Effects}} in {{Persuasion}}},
  shorttitle = {Argument {{Strength}} Is in the {{Eye}} of the {{Beholder}}},
  booktitle = {Proceedings of the 15th {{Conference}} of the {{European Chapter}} of the {{Association}} for {{Computational Linguistics}}: {{Volume}} 1, {{Long Papers}}},
  author = {Lukin, Stephanie and Anand, Pranav and Walker, Marilyn and Whittaker, Steve},
  editor = {Lapata, Mirella and Blunsom, Phil and Koller, Alexander},
  year = 2017,
  month = apr,
  pages = {742--753},
  publisher = {Association for Computational Linguistics},
  address = {Valencia, Spain},
  urldate = {2026-05-21}
}

@inproceedings{tan.etal_2016,
  title = {Winning {{Arguments}}: {{Interaction Dynamics}} and {{Persuasion Strategies}} in {{Good-faith Online Discussions}}},
  shorttitle = {Winning {{Arguments}}},
  booktitle = {Proceedings of the 25th {{International Conference}} on {{World Wide Web}}},
  author = {Tan, Chenhao and Niculae, Vlad and {Danescu-Niculescu-Mizil}, Cristian and Lee, Lillian},
  year = 2016,
  month = apr,
  series = {{{WWW}} '16},
  pages = {613--624},
  publisher = {International World Wide Web Conferences Steering Committee},
  address = {Republic and Canton of Geneva, CHE},
  doi = {10.1145/2872427.2883081},
  urldate = {2025-12-21},
  isbn = {978-1-4503-4143-1}
}

@inproceedings{lenz.bergmann_2024,
  ids = {lenz.bergmann_2024a},
  title = {{{PolArg}}: {{Unsupervised Polarity Prediction}} of~{{Arguments}} in~{{Real-Time Online Conversations}}},
  shorttitle = {{{PolArg}}},
  booktitle = {Robust {{Argumentation Machines}}},
  author = {Lenz, Mirko and Bergmann, Ralph},
  editor = {Cimiano, Philipp and Frank, Anette and Kohlhase, Michael and Stein, Benno},
  year = 2024,
  pages = {108--126},
  publisher = {Springer Nature Switzerland},
  address = {Cham},
  doi = {10.1007/978-3-031-63536-6_7},
  isbn = {978-3-031-63536-6},
  langid = {english}
}

@inproceedings{verma.etal_2026,
  title = {Predicting {{Convincingness}} in {{Political Speech}}: {{How Emotional Tone Shapes Persuasive Strength}}},
  shorttitle = {Predicting {{Convincingness}} in {{Political Speech}}},
  booktitle = {The {{Proceedings}} for the 15th {{Workshop}} on {{Computational Approaches}} to {{Subjectivity}}, {{Sentiment Social Media Analysis}} ({{WASSA}} 2026)},
  author = {Verma, Bhuvanesh and Marreddy, Mounika and Mehler, Alexander},
  editor = {Barnes, Jeremy and Barriere, Valentin and De Clercq, Orph{\'e}e and Klinger, Roman and Nouri, C{\'e}lia and Nozza, Debora and Singh, Pranaydeep},
  year = 2026,
  month = mar,
  pages = {37--51},
  publisher = {Association for Computational Linguistics},
  address = {Rabat, Morocco},
  urldate = {2026-03-25},
  isbn = {979-8-89176-378-4}
}

@article{falappa.etal_2011,
  title = {On the Evolving Relation between {{Belief Revision}} and {{Argumentation}}},
  author = {Falappa, Marcelo A. and Garc{\'i}a, Alejandro J. and {Kern-Isberner}, Gabriele and Simari, Guillermo R. and Falappa, Marcelo A. and Garc{\'i}a, Alejandro J. and {Kern-Isberner}, Gabriele and Simari, Guillermo R.},
  year = 2011,
  month = feb,
  journal = {The Knowledge Engineering Review},
  volume = {26},
  number = {1},
  pages = {35--43},
  issn = {1469-8005},
  doi = {10.1017/S0269888910000391},
  urldate = {2026-05-21},
  copyright = {2011 The Author(s)},
  langid = {english}
}

@inproceedings{habernal.gurevych_2016,
  ids = {habernalWhichArgumentMore2016},
  title = {Which Argument Is More Convincing? {{Analyzing}} and Predicting Convincingness of {{Web}} Arguments Using Bidirectional {{LSTM}}},
  shorttitle = {Which Argument Is More Convincing?},
  booktitle = {Proceedings of the 54th {{Annual Meeting}} of the {{Association}} for {{Computational Linguistics}} ({{Volume}} 1: {{Long Papers}})},
  author = {Habernal, Ivan and Gurevych, Iryna},
  year = 2016,
  pages = {1589--1599},
  publisher = {Association for Computational Linguistics},
  address = {Berlin, Germany},
  doi = {10.18653/v1/P16-1150},
  urldate = {2024-01-17},
  langid = {english}
}

@article{https://doi.org/10.1111/j.1740-9713.2004.00067.x,
author = {Wood, Michael},
title = {Statistical inference using bootstrap confidence intervals},
journal = {Significance},
volume = {1},
number = {4},
pages = {180-182},
doi = {https://doi.org/10.1111/j.1740-9713.2004.00067.x},
url = {https://rss.onlinelibrary.wiley.com/doi/abs/10.1111/j.1740-9713.2004.00067.x},
eprint = {https://rss.onlinelibrary.wiley.com/doi/pdf/10.1111/j.1740-9713.2004.00067.x},
year = {2004}
}

@misc{openai2025gptoss120bgptoss20bmodel,
      title={gpt-oss-120b \& gpt-oss-20b Model Card}, 
      author={OpenAI and Sandhini Agarwal and Lama Ahmad and Jason Ai and Sam Altman and Andy Applebaum and Edwin Arbus and Rahul K. Arora and Yu Bai and Bowen Baker and Haiming Bao and Boaz Barak and Ally Bennett and Tyler Bertao and Nivedita Brett and Eugene Brevdo and Greg Brockman and Sebastien Bubeck and Che Chang and Kai Chen and Mark Chen and Enoch Cheung and Aidan Clark and Dan Cook and Marat Dukhan and Casey Dvorak and Kevin Fives and Vlad Fomenko and Timur Garipov and Kristian Georgiev and Mia Glaese and Tarun Gogineni and Adam Goucher and Lukas Gross and Katia Gil Guzman and John Hallman and Jackie Hehir and Johannes Heidecke and Alec Helyar and Haitang Hu and Romain Huet and Jacob Huh and Saachi Jain and Zach Johnson and Chris Koch and Irina Kofman and Dominik Kundel and Jason Kwon and Volodymyr Kyrylov and Elaine Ya Le and Guillaume Leclerc and James Park Lennon and Scott Lessans and Mario Lezcano-Casado and Yuanzhi Li and Zhuohan Li and Ji Lin and Jordan Liss and Lily and Liu and Jiancheng Liu and Kevin Lu and Chris Lu and Zoran Martinovic and Lindsay McCallum and Josh McGrath and Scott McKinney and Aidan McLaughlin and Song Mei and Steve Mostovoy and Tong Mu and Gideon Myles and Alexander Neitz and Alex Nichol and Jakub Pachocki and Alex Paino and Dana Palmie and Ashley Pantuliano and Giambattista Parascandolo and Jongsoo Park and Leher Pathak and Carolina Paz and Ludovic Peran and Dmitry Pimenov and Michelle Pokrass and Elizabeth Proehl and Huida Qiu and Gaby Raila and Filippo Raso and Hongyu Ren and Kimmy Richardson and David Robinson and Bob Rotsted and Hadi Salman and Suvansh Sanjeev and Max Schwarzer and D. Sculley and Harshit Sikchi and Kendal Simon and Karan Singhal and Yang Song and Dane Stuckey and Zhiqing Sun and Philippe Tillet and Sam Toizer and Foivos Tsimpourlas and Nikhil Vyas and Eric Wallace and Xin Wang and Miles Wang and Olivia Watkins and Kevin Weil and Amy Wendling and Kevin Whinnery and Cedric Whitney and Hannah Wong and Lin Yang and Yu Yang and Michihiro Yasunaga and Kristen Ying and Wojciech Zaremba and Wenting Zhan and Cyril Zhang and Brian Zhang and Eddie Zhang and Shengjia Zhao},
      year={2025},
      eprint={2508.10925},
      archivePrefix={arXiv},
      primaryClass={cs.CL},
      url={https://arxiv.org/abs/2508.10925}, 
}

@misc{ollama2025,
  author       = {{Ollama}},
  title        = {Ollama},
  year         = {2025},
  howpublished = {\url{https://ollama.com}},
}

@inproceedings{reimers-2019-sentence-bert,
  title = {Sentence-{{BERT}}: {{Sentence}} Embeddings Using Siamese {{BERT-networks}}},
  booktitle = {Proceedings of the 2019 Conference on Empirical Methods in Natural Language Processing},
  author = {Reimers, Nils and Gurevych, Iryna},
  year = 2019,
  month = nov,
  publisher = {Association for Computational Linguistics}
}

@article{liu.etal_2024a,
  title = {Lost in the {{Middle}}: {{How Language Models Use Long Contexts}}},
  shorttitle = {Lost in the {{Middle}}},
  author = {Liu, Nelson F. and Lin, Kevin and Hewitt, John and Paranjape, Ashwin and Bevilacqua, Michele and Petroni, Fabio and Liang, Percy},
  year = 2024,
  journal = {Transactions of the Association for Computational Linguistics},
  volume = {12},
  pages = {157--173},
  publisher = {MIT Press},
  address = {Cambridge, MA},
  doi = {10.1162/tacl_a_00638},
  urldate = {2026-05-21}
}

@inproceedings{page.etal_1999,
  title = {The {{PageRank Citation Ranking}} : {{Bringing Order}} to the {{Web}}},
  shorttitle = {The {{PageRank Citation Ranking}}},
  booktitle = {The {{Web Conference}}},
  author = {Page, Lawrence and Brin, Sergey and Motwani, R. and Winograd, T.},
  year = 1999,
  month = nov,
  urldate = {2026-05-21}
}

@InProceedings{gienapp:2020,
    author =              {Gienapp, Lukas and Stein, Benno and Hagen, Matthias and Potthast, Martin},
    booktitle =           {The 58th annual meeting of the Association for Computational Linguistics (ACL) },
    month =               jul,
    publisher =           {ACL},
    site =                {Seattle, USA},
    title =               {{Efficient Pairwise Annotation of Argument Quality}},
    year =                2020
}

@misc{he2021debertav3,
      title={DeBERTaV3: Improving DeBERTa using ELECTRA-Style Pre-Training with Gradient-Disentangled Embedding Sharing}, 
      author={Pengcheng He and Jianfeng Gao and Weizhu Chen},
      year={2021},
      eprint={2111.09543},
      archivePrefix={arXiv},
      primaryClass={cs.CL}
}

@inproceedings{amgoud.ben-naim_2018a,
  title = {Weighted {{Bipolar Argumentation Graphs}}: {{Axioms}} and {{Semantics}}},
  shorttitle = {Weighted {{Bipolar Argumentation Graphs}}},
  booktitle = {Proceedings of the {{Twenty-Seventh International Joint Conference}} on {{Artificial Intelligence}}},
  author = {Amgoud, Leila and {Ben-Naim}, Jonathan},
  year = 2018,
  month = jul,
  pages = {5194--5198},
  publisher = {International Joint Conferences on Artificial Intelligence Organization},
  address = {Stockholm, Sweden},
  doi = {10.24963/ijcai.2018/720},
  urldate = {2026-05-21},
  isbn = {978-0-9992411-2-7},
  langid = {english}
}

@article{baroni.etal_2019,
  title = {From Fine-Grained Properties to Broad Principles for Gradual Argumentation: {{A}} Principled Spectrum},
  shorttitle = {From Fine-Grained Properties to Broad Principles for Gradual Argumentation},
  author = {Baroni, Pietro and Rago, Antonio and Toni, Francesca},
  year = 2019,
  month = feb,
  journal = {International Journal of Approximate Reasoning},
  volume = {105},
  pages = {252--286},
  issn = {0888-613X},
  doi = {10.1016/j.ijar.2018.11.019},
  urldate = {2026-05-21}
}

@article{lippi.torroni_2016,
  ids = {lippiArgumentationMiningState2016},
  title = {Argumentation {{Mining}}: {{State}} of the {{Art}} and {{Emerging Trends}}},
  shorttitle = {Argumentation {{Mining}}},
  author = {Lippi, Marco and Torroni, Paolo},
  year = 2016,
  month = apr,
  journal = {ACM Transactions on Internet Technology},
  volume = {16},
  number = {2},
  pages = {1--25},
  issn = {1533-5399, 1557-6051},
  doi = {10.1145/2850417},
  urldate = {2024-01-17},
  langid = {english}
}

@book{baroniHandbookFormalArgumentation2018,
  ids = {baroni.etal_2018},
  title = {Handbook of Formal Argumentation},
  editor = {Baroni, Pietro and Gabbay, Dov M. and Giacomin, Massimiliano and {vander Torre}, Leendert},
  year = 2018,
  publisher = {College Publications},
  isbn = {978-1-84890-275-6},
  langid = {english}
}

@article{dung_1995,
  title = {On the Acceptability of Arguments and Its Fundamental Role in Nonmonotonic Reasoning, Logic Programming and {\emph{n}}-Person Games},
  author = {Dung, Phan Minh},
  year = 1995,
  month = sep,
  journal = {Artificial Intelligence},
  volume = {77},
  number = {2},
  pages = {321--357},
  issn = {0004-3702},
  doi = {10.1016/0004-3702(94)00041-X},
  urldate = {2026-05-20}
}

@book{gabbay.etal_2021,
  title = {Handbook of {{Formal Argumentation}}, {{Volume}} 2},
  editor = {Gabbay, Dov and Giacomin, Massimiliano and Guillermo, Simari R. and Thimm, Matthias},
  year = 2021,
  month = aug,
  volume = {2},
  urldate = {2025-02-13},
  isbn = {978-1-84890-336-4}
}

@article{budzynska.etal_2016,
  title = {Theoretical Foundations for Illocutionary Structure Parsing},
  author = {Budzynska, Katarzyna and Janier, Mathilde and Reed, Chris and {Saint-Dizier}, Patrick},
  year = 2016,
  month = jan,
  journal = {Argument \& Computation},
  volume = {7},
  number = {1},
  pages = {91--108},
  publisher = {IOS Press},
  issn = {1946-2166},
  doi = {10.3233/AAC-160005},
  urldate = {2024-10-02},
  langid = {english}
}

@inproceedings{cabrioFiveYearsArgument2018,
  ids = {cabrio.villata_2018},
  title = {Five {{Years}} of {{Argument Mining}}: {{A Data-driven Analysis}}},
  shorttitle = {Five {{Years}} of {{Argument Mining}}},
  booktitle = {Proceedings of the {{Twenty-Seventh International Joint Conference}} on {{Artificial Intelligence}}},
  author = {Cabrio, Elena and Villata, Serena},
  year = 2018,
  month = jul,
  pages = {5427--5433},
  publisher = {International Joint Conferences on Artificial Intelligence Organization},
  address = {Stockholm, Sweden},
  doi = {10.24963/ijcai.2018/766},
  urldate = {2024-01-17},
  eventtitle = {Twenty-{{Seventh International Joint Conference}} on {{Artificial Intelligence}} \textbraceleft{{IJCAI-18}}\textbraceright},
  isbn = {978-0-9992411-2-7},
  langid = {english}
}

@article{modgil.prakken_2014,
  title = {The {{ASPIC}}+ Framework for Structured Argumentation: A Tutorial},
  shorttitle = {The {{ASPIC}}+ Framework for Structured Argumentation},
  author = {Modgil, Sanjay and Prakken, Henry},
  year = 2014,
  month = jan,
  journal = {Argument \& Computation},
  volume = {5},
  number = {1},
  pages = {31--62},
  publisher = {Taylor \& Francis},
  issn = {1946-2166},
  doi = {10.1080/19462166.2013.869766},
  urldate = {2024-05-29}
}

@article{garcez.lamb_2023,
  title = {Neurosymbolic {{AI}}: The 3rd Wave},
  shorttitle = {Neurosymbolic {{AI}}},
  author = {d'Avila Garcez, Artur and Lamb, Lu{\'i}s C.},
  year = 2023,
  month = nov,
  journal = {Artificial Intelligence Review},
  volume = {56},
  number = {11},
  pages = {12387--12406},
  issn = {1573-7462},
  doi = {10.1007/s10462-023-10448-w},
  urldate = {2026-05-20},
  langid = {english}
}

@article{ruiz-dolz.etal_2022,
  title = {Automatic {{Debate Evaluation}} with {{Argumentation Semantics}} and {{Natural Language Argument Graph Networks}}},
  author = {{Ruiz-Dolz}, Ramon and Heras, Stella and {Garc{\'i}a-Fornes}, Ana},
  year = 2022,
  publisher = {arXiv},
  doi = {10.48550/ARXIV.2203.14647},
  urldate = {2024-07-01},
  copyright = {Creative Commons Attribution Non Commercial Share Alike 4.0 International}
}

@article{shi.etal_2026,
  ids = {shi.etal_2026a},
  title = {From Text Mining to Intelligent Debate: {{Task}} Frameworks and Technological Evolution in Computational Argumentation},
  shorttitle = {From Text Mining to Intelligent Debate},
  author = {Shi, Lida and Giunchiglia, Fausto and Wang, Hanwen and Cheng, Yongqi and Song, Rui and Shi, Daqian and Diao, Xiaolei and Xu, Hao},
  year = 2026,
  month = mar,
  journal = {Information Processing \& Management},
  volume = {63},
  number = {2, Part B},
  pages = {104465},
  issn = {0306-4573},
  doi = {10.1016/j.ipm.2025.104465},
  urldate = {2025-12-17}
}

@inproceedings{morio.fujita_2018b,
  title = {Annotating {{Online Civic Discussion Threads}} for {{Argument Mining}}},
  booktitle = {2018 {{IEEE}}/{{WIC}}/{{ACM International Conference}} on {{Web Intelligence}} ({{WI}})},
  author = {Morio, Gaku and Fujita, Katsuhide},
  year = 2018,
  month = dec,
  pages = {546--553},
  doi = {10.1109/WI.2018.00-39},
  urldate = {2025-02-10}
}

@article{visserArgumentation2016US2020,
  ids = {visser.etal_2020},
  title = {Argumentation in the 2016 {{US}} Presidential Elections: {{Annotated}} Corpora of Television Debates and Social Media Reaction},
  shorttitle = {Argumentation in the 2016 {{US}} Presidential Elections},
  author = {Visser, Jacky and Konat, Barbara and Duthie, Rory and Koszowy, Marcin and Budzynska, Katarzyna and Reed, Chris},
  year = 2020,
  month = mar,
  journal = {Language Resources and Evaluation},
  volume = {54},
  number = {1},
  pages = {123--154},
  issn = {1574-020X, 1574-0218},
  doi = {10.1007/s10579-019-09446-8},
  urldate = {2024-01-17},
  langid = {english}
}

\end{document}